%% file: main.tex
\documentclass[10pt,twocolumn,letterpaper]{article}

\usepackage{cvpr}              % To produce the CAMERA-READY version
\input{preamble}

\definecolor{cvprblue}{rgb}{0.21,0.49,0.74}
\usepackage[pagebackref,breaklinks,colorlinks,citecolor=cvprblue]{hyperref}
\usepackage[accsupp]{axessibility}
\usepackage{abstract}
\def\paperID{1503} % *** Enter the Paper ID here
\def\confName{CVPR}
\def\confYear{2024}

\title{Human Gaussian Splatting: Real-time Rendering of Animatable Avatars}

\author{Arthur Moreau\thanks{Authors contributed equally to this work.} \qquad\quad Jifei Song\footnotemark[1] \qquad\quad  Helisa Dhamo \qquad\quad Richard Shaw\\Yiren Zhou \qquad\quad Eduardo Pérez-Pellitero\\
Huawei Noah's Ark Lab
}
\usepackage{multirow}
\newcommand{\OURS}{\mbox{HuGS}}

\begin{document}

\twocolumn[{
\maketitle
    \begin{center}
        \captionsetup{type=figure}
        %\vspace{-1.2em}
        \includegraphics[width=\linewidth]{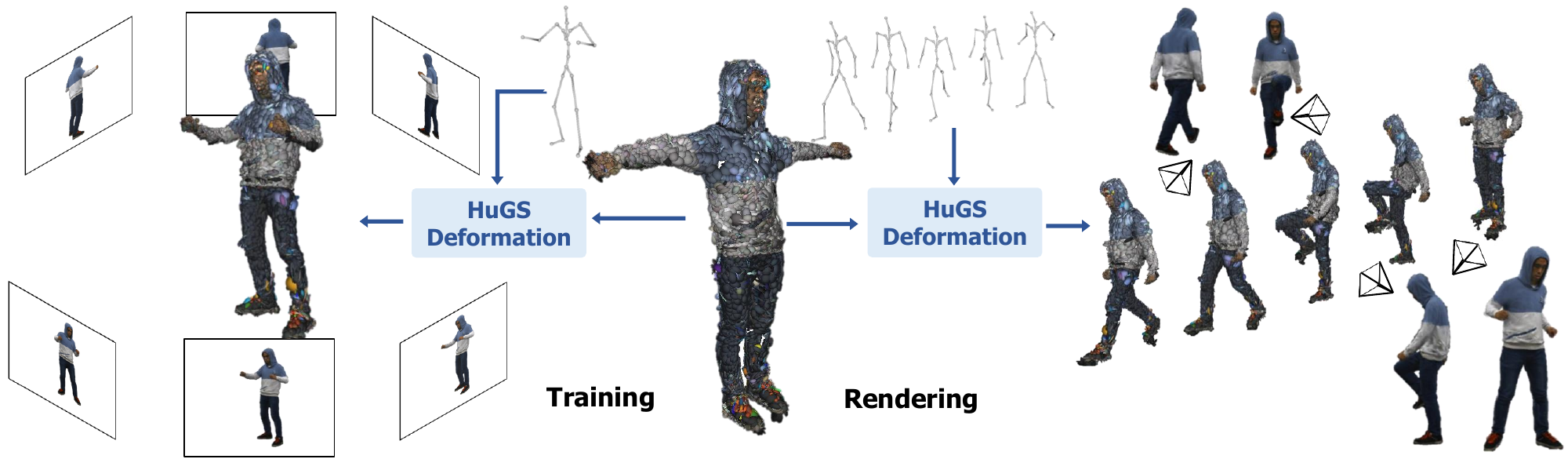}
        %\vspace{-1.7em}
        %\vspace{-0.5cm}
        \caption{\textbf{Overview of \OURS.} Using multi-view video frames of a dynamic human, {\OURS} learns a photorealistic 3D human avatar represented by a 3D Gaussian Splatting model. Given an arbitrary human body pose, our model deforms the canonical 3D representation to the observation space, from which novel views can be rendered in real-time from any camera viewpoint. }
       \label{fig:teaser_figure}
    \end{center}
}]

\thispagestyle{empty}
\input{sec/0_abstract}    
\saythanks
\input{sec/1_intro}
\input{sec/2_related_work}
\input{sec/3_method}
\input{sec/4_experiments}
\input{sec/5_discussion}
\input{sec/6_conclusion}

\clearpage
{
    \small
    \bibliographystyle{ieeenat_fullname}
    \bibliography{main}
}

\input{sec/X_suppl}

\end{document}

%% file: preamble.tex
\usepackage[dvipsnames]{xcolor}
\usepackage{colortbl}
\usepackage{float}
\definecolor{orange}{rgb}{0.94,0.67,0.36}

%% file: sec/0_abstract.tex
\begin{abstract}
\it \normalsize
This work addresses the problem of real-time rendering of photorealistic human body avatars learned from multi-view videos. While the classical approaches to model and render virtual humans generally use a textured mesh, recent research has developed neural body representations that achieve impressive visual quality. However, these models are difficult to render in real-time and their quality degrades when the character is animated with body poses different than the training observations. We propose an animatable human model based on 3D Gaussian Splatting, that has recently emerged as a very efficient alternative to neural radiance fields. The body is represented by a set of gaussian primitives in a canonical space which is deformed with a coarse to fine approach that combines forward skinning and local non-rigid refinement. We describe how to learn our Human Gaussian Splatting (\OURS) model in an end-to-end fashion from multi-view observations, and evaluate it against the state-of-the-art approaches for novel pose synthesis of clothed body. Our method achieves $1.5$ dB PSNR  improvement over the state-of-the-art on THuman4 dataset while being able to render in real-time ($\approx$ $80$ fps for $512 \times 512$ resolution).
\end{abstract}

%% file: sec/1_intro.tex
\section{Introduction}
\label{sec:intro}

Virtual human avatars are essential components of virtual reality and video games for applications such as content creation and immersive interaction between users and virtual worlds. The common procedure to create a highly realistic avatar of a person involves expensive sensors and tedious manual work. However, recent progress in 3D modeling and neural rendering enabled data-driven models that learn controllable human avatars from images. 

Recent research in this area focuses on neural representations based on Neural Radiance Fields (NeRF)~\cite{mildenhall2020nerf} or Signed Distance Fields (SDF)~\cite{park2019deepsdf}. Thanks to their continuous design, they can represent highly detailed shape and textures of a clothed human body, and thus exhibit better quality than the commonly-used textured meshes. However, their deployment is difficult, first because they present long training and rendering times, but also because animating the character in a controllable way is challenging. As a result, these neural avatars shine on novel view synthesis of body poses observed during training, but struggle to generalize to novel body poses with similar quality.

While animating meshes is a well understood problem that has been around for decades, \eg using direct skinning algorithms~\cite{Lewis2000PoseSD, merry2006animation, kavan2007skinning}, transferring these ideas for implicit neural representations is challenging. Some efforts have been made to learn skinning weights fields, but they often define a backward skinning methodology~\cite{peng2021animatable,ARAH:ECCV:2022, li2022tava}, which is pose-dependent and where solutions are slow and do not generalize well to unseen body poses.

In this work, we represent the human body with 3D Gaussian Splatting (3D-GS)~\citep{kerbl3Dgaussians}. This novel paradigm for view synthesis uses an explicit set of primitives shaped as 3D Gaussians to represent the radiance field of a scene. This enables fast tile-based rasterization, which is orders of magnitudes faster than the rendering speed of implicit methods based on ray marching. Leveraging its explicit and discrete nature, we explore its capability to be deformed with direct forward skinning, similar to a mesh. We observe that using Linear Blend Skinning (LBS) with skinning weights learned for each Gaussian provides an efficient and effective method that generalizes well to new body poses, but is not expressive enough to capture the local garment deformations of clothed avatars. We propose to refine the deformation of Gaussians with a shallow neural network that captures the local movements of the body surface.
 
The proposed algorithm, named \textbf{Human Gaussian Splatting (HuGS)}, paves the way for animatable human body models based on Gaussian Splatting. Our contributions can be summarized as follows:

\begin{enumerate}
    \item We propose a first algorithm for novel pose synthesis of the human body based on 3D Gaussian Splatting.
    \item We define a coarse-to-fine approach to animate the set of Gaussian primitives, based on forward skinning for skeleton-based movements and a pose-dependent MLP for local garments deformations.
    \item We compare HuGS to neural-based human models on three public datasets and exhibit results on-par or better than state-of-the-art methods, while rendering one order of magnitude faster.
\end{enumerate}

%% file: sec/2_related_work.tex
\section{Related Work}
\label{sec:related_work}

%-------------------------------------------------------------------------
\paragraph{Differentiable rendering of radiance fields}

Learning 3D representations from 2D images for novel view synthesis has been very active in the last few years~\citep{mittal2023neural, nerf_review, jang2023vschh} since the seminal work of Neural Radiance Fields~\cite{mildenhall2020nerf}. While NeRF methods usually define emitted color and density of each point of a static scene, it has been adapted to model dynamic content by incorporating the time dimension in the representation~\cite{pumarola2020d, li2020neural,park2021nerfies}. These dynamic radiance fields can be used to train volumetric representation of humans in movement~\cite{peng2023representing,isik2023humanrf} and to replay an existing video from a new camera viewpoint. NeRF and related approaches rely on ray marching and volumetric rendering~\cite{tagliasacchi2022volume}, requiring to evaluate a neural network on many points along each camera ray. Making this evaluation more efficient can be tackled by storing information in an explicit way~\citep{yu_and_fridovichkeil2021plenoxels, SunSC22, xu2022point, li2023read, kulhanek2023tetranerf, mueller2022instant, barron2023zipnerf} but the ray marching design inherently limits the rendering time of these methods. 

By contrast, 3D Gaussian Splatting~\cite{kerbl3Dgaussians} models the radiance field of a scene with explicit primitives shaped as 3D gaussians. The main advantage comes from the fast rendering step that avoids ray marching by sorting and splatting primitives w.r.t.\ the camera position, enabling real-time applications.  Each primitive is defined by its position, covariance matrix, view-dependent color represented with spherical harmonics, and opacity. These parameters are optimized through gradient descent to reconstruct the observed images with high fidelity. Recent works have extended 3D-GS to dynamic scenes by learning time-dependent gaussian parameters~\cite{luiten2023dynamic, yang2023deformable3dgs, wu20234dgaussians, yang2023gs4d, shaw2023swags}. Instead, our work focuses on a model tailored for a drivable human by learning gaussians deformations that depend on the body pose.
%-------------------------------------------------------------------------

%-------------------------------------------------------------------------
\paragraph{Animatable human body models}

We refer to an \textit{animatable body model} as a 3D representation of a human character, which is defined in a canonical space (i.e.\ a reference body pose, often set as T-pose) and can be deformed to represent any body pose in their respective observation (or posed) space. Pioneer works in human body modelling are statistical mesh templates such as SMPL~\cite{SMPL:2015,SMPL-X:2019}. Fitted to 3D scans of a large set of people, they provide a parametric representation of the human body shape. The deformation from canonical to observation space can be computed easily with linear blend skinning and thus these templates are often used as a building block of more complex methods. Avatars can be learned from different data modalities, such 3D scans~\cite{chen2021snarf, Saito:CVPR:2021, su2023caphy}, monocular videos~\cite{weng_humannerf_2022_cvpr, yu2023monohuman} or a single image~\cite{liao2023car, caselles2023sira}. We present below methods that use multiview RGB video capture.

Recent literature has explored the use of neural representations, either based on NeRF or SDF. These methods perform ray marching in the observation space but usually define a static radiance field in the canonical space. One approach is to establish backward correspondences (from observation to canonical) for each point. Neural Actor~\citep{liu2021neural}, Animatable NeRF~\citep{peng2021animatable} and InstantNVR~\citep{instant_nvr} use pose-dependent networks to learn deformation or blend-weights fields, but they have been observed to generalize poorly to novel body poses. ARAH~\citep{ARAH:ECCV:2022}, TAVA~\citep{li2022tava} and PoseVocab~\citep{li2023posevocab} rather use a joint root-finding approach that generalizes better but is slow to compute. 

Another line of work, closer to ours, circumvent the backward correspondence problem by applying forward skinning deformation on different kind of canonical \textit{primitives}, resulting in a radiance field defined in the observation space. Neural Body~\citep{peng2021neural} anchors latent codes to SMPL vertices which are then decoded to density and colors by a CNN. SLRF~\citep{SLRF} and AvatarRex~\citep{zheng2023avatarrex} use multiple local NeRFs centered on vertices whose origins are moved with LBS. DVA~\citep{remelli2022drivable} computes deformation of articulated volumetric primitives~\citep{Lombardi21}. Several methods use point-based primitives, for which forward skinning is well defined, either with volumetric rendering~\citep{su2023npc, yu2023point} or rasterization~\citep{Zheng2023pointavatar, prokudin2023dynamic} approaches. Overall, forward skinning is more convenient to use than backward approaches. However, by defining the primitives on the SMPL template surface, such approaches often struggle to represent characters with loose clothing.

Finally, skeleton-based rigid deformations are not a sufficient driving signal to explain all the movement observed in the data~\citep{bagautdinov2021driving}. Local garment wrinkles or muscles contractions are typical examples of non-rigid motion that need to be addressed. Modelling these phenomenons in a proper way would require physics-based modelling~\citep{su2023caphy}, but this remains difficult together with RGB supervision. Most learning-based methods rather use local refinement of the geometry with pose-dependent neural networks~\citep{SLRF, li2023posevocab,li2022tava}. Some methods also add local shading components~\cite{li2022tava,bagautdinov2021driving} to approximate the ambient occlusion used in graphics pipelines.

%% file: sec/3_method.tex
\section{Method}
\label{sec:method}

\begin{figure*}[t]
   \centering
   \includegraphics[width=\linewidth]{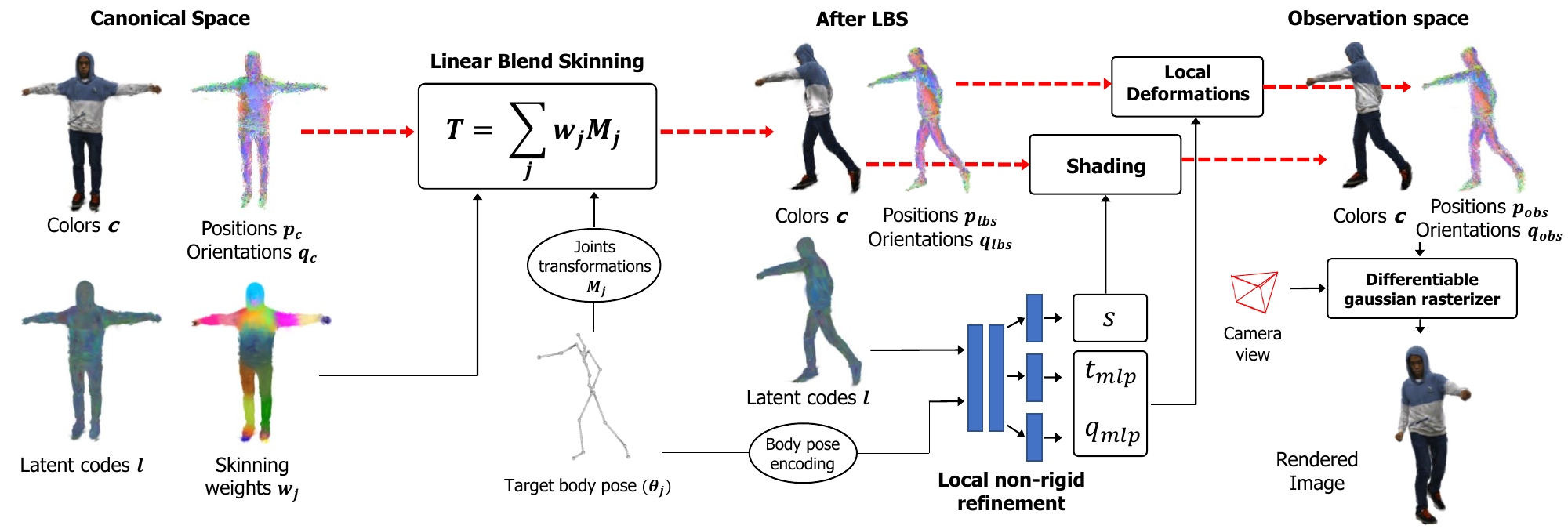}
   \caption{\textbf{{\OURS} method overview.} We represent the different attributes of Gaussians at each step of the deformation pipeline. Canonical positions and orientations are first deformed with LBS using the learned skinning weights. Then positions, orientations, and colors are refined by an MLP using the latent codes. Finally, Gaussians in the observation space are rendered through the target camera view.}
   \label{fig:method_figure}
\end{figure*}

Our goal is to reconstruct human avatars from multi-view videos and render images of the reconstructed virtual character from arbitrary camera view and body poses, \ie, novel pose synthesis. An overview of our method is presented in Figure~\ref{fig:method_figure}. We train our model from a collection of multiview videos depicting various body poses, captured by $N$ calibrated cameras during $T$ timesteps. We pre-compute or assume access to SMPL~\cite{SMPL:2015} or SMPL-X~\cite{SMPL-X:2019} parameters, \ie, body shape $\boldsymbol{\beta}$ and body poses $\boldsymbol{\theta_t}=(\theta_1,\theta_j,\ldots,\theta_J)$ expressed as the 3D rotation of each body joint $j$, at timestep $t$. We also use foreground segmentation masks.

%-------------------------------------------------------------------------
\subsection{Canonical representation}
\label{sec:body_representation}

We represent the canonical human body as a set of volumetric primitives shaped as 3D Gaussians. Each Gaussian is parametrized by its own set of learnable parameters.
\begin{itemize}
    \item a 3D canonical position $\mathbf{p}_c = (p_x,p_y,p_z)$,
    \item a 3D orientation, represented by a quaternion $\mathbf{q}_c$,
    \item a 3D scale $\mathbf{s}=(s_{x},s_{y},s_{z})$,
    \item a color $\mathbf{c}=(c_r,c_g,c_b)$,
    \item an opacity scalar value $o$,
    \item a skinning weight vector $\mathbf{w} = (w_1,w_j,\ldots,w_{J})$ that regulates the influence of each body joint $j$ on how the gaussian moves,
    \item a latent code $\mathbf{l}$ that encodes the non-rigid motion.
\end{itemize}

This builds up from the original 3D Gaussian splatting formulation~\cite{kerbl3Dgaussians}, with the addition of the last 2 parameters that encode the pose-dependent movement of each primitive. We consider scale and opacity consistent across novel views and novel poses. For a given target body pose, we transform canonical position, orientation, and color to the posed space, as described in sections~\ref{sec:lbs} and ~\ref{sec:mlp}. 

%-------------------------------------------------------------------------
\subsection{Deformation with forward skinning.}
\label{sec:lbs}

We use Linear Blend Skinning (LBS) to deform our model. We consider body joints in the canonical space imported from the SMPL template model. Given a body pose $\boldsymbol{\theta_{t}}$, we can compute the rigid transformations $\mathbf{M_{j}} \in \mathrm{SE(3)}$ for the $j$-th body joint using the kinematic tree. Then, each gaussian's skinning transformation $\mathbf{T_{t}}$ for pose $\boldsymbol{\theta_{t}}$ is defined by weight-averaging joint transformations according to skinning weights $\mathbf{w}$:
\begin{equation}
\mathbf{T_{t}} = \sum_{j=1}^{J} w_{j} \mathbf{M_{j}}.
\end{equation}

The canonical Gaussian position is then transformed to the posed space using $\mathbf{T_{t}}$. We also rotate the gaussian using the rotation component $\mathbf{R_{t}}$ of $\mathbf{T_{t}} = [\mathbf{R_{t}} | \mathbf{t_{t}}]$:

\begin{equation}
\mathbf{p}_{lbs} = \mathbf{T_{t}} \mathbf{p}_c \qquad \mathbf{q_{lbs}} = \mathbf{R_{t}} \circ \mathbf{q}_{c}.
\end{equation}

We apply directly forward skinning on the canonical primitives, similar to mesh deformation in common pipelines, and learn only the skinning weights attached to each Gaussian. In contrast, previous NeRF and SDF approaches, because they rely on ray marching on the posed space, need backward skinning formulations~\cite{peng2021animatable,li2022tava, ARAH:ECCV:2022} which is notoriously more difficult and/or slower.

\subsection{Local non-rigid refinement}
\label{sec:mlp}

LBS moves the Gaussians towards the target body pose and provides excellent generalization to novel poses, but only encodes the rigid deformations of the body joints. Because we want our method to be able to operate on clothed avatars, we also need to model local non-rigid deformations caused by garments. To this end, we compute per-gaussian residual outputs learned by a pose-dependent MLP that can translate, rotate, and change the lightness of the primitive.

\paragraph{Body pose encoding} We want our model to learn how Gaussians move w.r.t.\ the body pose rather than w.r.t.\ time, but we also expect the network to learn \textit{local} deformations. Thus, using the global pose vector $\boldsymbol{\theta_{t}}$ as input can represent too many information for local primitives whose deformation depends on a nearby joints orientations only. This can ultimately enable the network to learn spurious correlations and overfit~\cite{li2023posevocab}. Following SCANimate~\cite{Saito:CVPR:2021}, we use an attention-weighting scheme which uses the skinning weights $\mathbf{w}$ and an explicit attention map $W$ based on the kinematic tree to define the local pose vector $\boldsymbol\theta^{l}_{t}$:
\begin{equation}
\boldsymbol{\theta^{l}_{t}} = (W \cdot \mathbf{w}) \odot \boldsymbol{\theta_{t}},
\end{equation}

where $ \boldsymbol{\theta_{t}}$ is represented as a vector of quaternions and $\odot$ denotes element-wise multiplication.

\paragraph{Shading} While our approach does not model view-dependent specularities because we consider the human body as Lambertian, we allow the color to change depending on the local body pose. This enables us to take into account self-occlusions and shadows caused by garment wrinkles. This shading approximates ambient occlusion in graphics pipelines~\cite{mendez2009obscurances}. Formally, the MLP outputs a scaling factor $s \in [0,2]$ that multiplies the RGB color of each Gaussian, which is then clipped in $[0,1]$. 

\paragraph{MLP architecture} Our neural network takes as input the local body pose vector $ \boldsymbol{\theta_{t}}$ concatenated with the per-gaussian learnable latent features $\mathbf{l}$. This latent code identifies each primitive and encodes their local motion depending on the body pose which is then decoded by the MLP. We show in Section~\ref{sec:ablations} that using this latent code leads to better performance than using the position. This information is processed by 2 hidden layers with 64 neurons and ReLU activations and then decoded by 3 separate heads with 2 layers that output respectively a translation vector $\mathbf{t}_{\text{mlp}}$, a quaternion that represents how the Gaussian rotates $\mathbf{q}_{\text{mlp}}$ and the ambient occlusion scaling factor $s$. 

We obtain the final Gaussian position $p_{\text{obs}}$ and orientation $\mathbf{q}_{\text{obs}}$ by applying this residual transformation to the LBS output. Importantly, to enable generalization the residual translation vector is defined in the canonical space and thus needs to be rotated with the orientation from LBS:

\begin{equation}
\mathbf{p}_{\text{obs}} = \mathbf{p}_{\text{lbs}} + (\mathbf{R}_{t}\mathbf{t}_{\text{mlp}}) \qquad \mathbf{q}_{\text{obs}} = \mathbf{q}_{\text{mlp}} \circ \mathbf{q}_{\text{lbs}}.
\end{equation}

\subsection{Image rendering}

Once the parameters of Gaussians in the observation space have been computed, we render the image using the fast and differentiable Gaussian rasterizer from 3D-GS~\cite{kerbl3Dgaussians}.

\subsection{Training procedure}
\label{sec:training_objective}

At each training step, given an image and its corresponding body pose, we transform the canonical Gaussians from the canonical space to the observation space, render the image and finally optimize parameters from the Gaussians and the MLP with gradient descent. We use the SMPL model to initialize the primitives. Gaussian centers are set to the vertices positions, from which we can import the skinning weights from the template. During training, the set of Gaussians is incrementally densified and pruned, following the heuristics proposed by 3D-GS.
We describe below the optimization objective of our method, which combines reconstruction and regularization losses. As shown in section~\ref{sec:ablations}, regularization plays an important role in guiding our overparametrized model to a solution that generalizes to novel body poses.

\paragraph{Reconstruction losses} The main objective of the model is to reconstruct the training images. Using the segmentation mask, we set the background pixels from the ground-truth image in black. We use a $L_1$ loss $\mathcal{L}_{L_1} $, a D-SSIM~\cite{wang2004image} loss $\mathcal{L}_{\text{ssim}}$ and a perceptual loss~\cite{zhang2018unreasonable} $\mathcal{L}_{\text{lpips}}$ with VGG~\cite{simonyan2014very} weights between rendered and groundtruth images.

\begin{figure}[t]
   \centering
   \includegraphics[width=\linewidth]{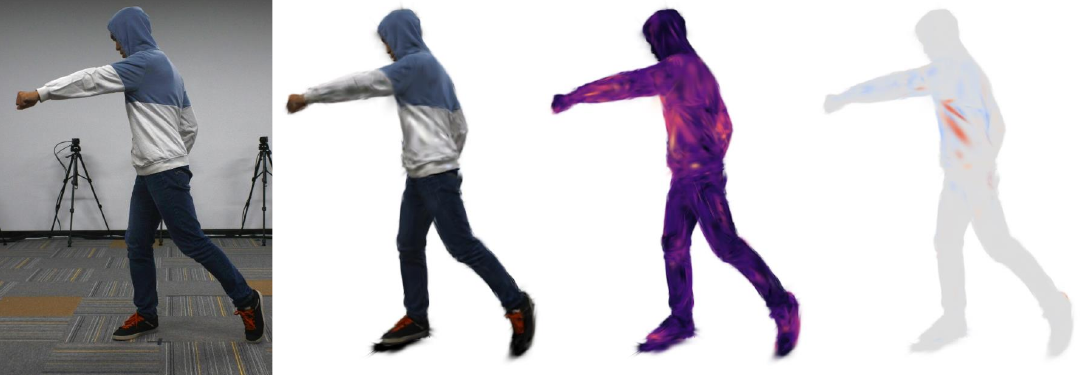}
   \caption{\textbf{Visualization of MLP outputs.} From left to right: ground-truth image, rendered image, translation output $t_{\text{mlp}}$ norm (lightest colors indicate largest translation vector) and ambient occlusion factor $s$ (grey: no color modification, blue: darker colors, red: lighter color). We observe that our MLP learns to operate on the dynamic parts of garments.}
   \label{fig:mlp_figure}
\end{figure}

\paragraph{Minimize MLP output:} We want our deformations to rely on LBS as much as possible and expect the MLP to learn only local deformations. Thus, we restrict the MLP outputs to be as small as possible. We define one regularization term for each output: $\mathcal{L}_{\text{trans}}$ controls the norm of the translation residual, $\mathcal{L}_{\text{rot}}$ pushes the rotation residual close to the identity quaternion $q_{\text{id}}$, and $\mathcal{L}_{\text{amb}}$ guides the ambient occlusion factor to stay close to 1:

%\begin{equation}
$
\mathcal{L}_{\text{trans}} = \| t_{\text{mlp}} \|_{2} \quad \mathcal{L}_{\text{rot}} = \| q_{\text{mlp}} - q_{\text{id}} \|_{1} \quad \mathcal{L}_{s} = \| s - 1 \|_{2}
$.
%\end{equation}

%\begin{equation}
%\mathbf{\mathcal{L}_{trans}} = \| x_{mlp} \|_{2} \quad \mathbf{\mathcal{L}_{rot}} = g(q_{mlp}) \quad \mathbf{\mathcal{L}_{s}} = \| s - 1 \|_{2}
%\end{equation}

%Where $g(R) = cos^{-1}(\frac{tr(R)-1}{2})$ is the geodesic distance between a rotation matrix $r$ and the identity.

\paragraph{Regularization of canonical positions:} We encourage Gaussian positions to stay close to the SMPL mesh to avoid floating artifacts. Because our model represents the clothed body, we define a threshold $\tau_{\text{pos}}$ that represents the maximum distance between the skin and the clothes. We search the nearest vertex $v_{i}$ from each Gaussian canonical position $p_{i}$ and apply the following loss, that penalizes points that are further than the threshold:
%\begin{equation}
 \begin{center}
 $\mathcal{L}_{\text{mesh}} = \sum_{i} ReLU(\| v_{i} - p_{i} \|_{2} - \tau_{\text{pos}} )$.
 \end{center}
%\end{equation}

\paragraph{Skinning weights weak supervision:} Because we train on a limited amount of gestures, the training data can usually be fitted with skinning weights that do not generalize well. Consequently, we softly supervise the skinning weights of our Gaussians with those from SMPL, \ie, the closer a Gaussian is from a vertex, the more similar its skinning weights need to be: 

%\begin{equation}
 \begin{center}
 $\mathcal{L}_{\text{skn}} = \sum_{i} {\text{ReLU}}(\| w(g_{i}) - w(v_{i}) \|_{2} - \tau_{\text{skn}} \| v_{i} - p_{i} \|_{2} )$.
 \end{center}
%\end{equation}

It should be noted that we do not backpropagate the gradient of the distance $\| v_{i} - p_{i} \|_{2}$ through this loss.

\vspace{0.5cm}
We sum all the reconstruction and regularization terms to obtain the final loss $\mathbf{\mathcal{L}}$. The weights and hyperparameters used are given in the supplementary materials.

%\begin{multline}
%\mathbf{\mathcal{L}} = \lambda_{L_1}\mathcal{L}_{L_1} + \lambda_{\text{ssim}}\mathcal{L}_{\text{ssim}} +  \lambda_{\text{lpips}}\mathcal{L}_{\text{lpips}} + \lambda_{\text{trans}}\mathcal{L}_{\text{trans}} \\
%+ \lambda_{\text{rot}}\mathcal{L}_{\text{rot}} + \lambda_{s}\mathcal{L}_{s} + \lambda_{\text{mesh}}\mathcal{L}_{\text{mesh}} + \lambda_{\text{skn}}\mathcal{L}_{\text{skn}}
%\end{multline}

%-------------------------------------------------------------------------

%% file: sec/4_experiments.tex
\section{Experiments}
\label{sec:experiments}

\input{tables/thuman_full}

In Section~\ref{sec:comparison}, we compare Human Gaussian Splatting against state-of-the-art human avatars on novel views and novel pose synthesis on public datasets. Then, we perform several ablation studies in Section~\ref{sec:ablations} to show the benefit our design choices. This research has been conducted with public datasets only, which, to the best of our knowledge, have collected human subject data according to regulations. More results are shown in supplementary materials, including videos of novel pose synthesis and a demonstration of real-time rendering.

\subsection{Implementation details}
Our algorithm is implemented in the PyTorch framework. The LBS module is inspired by the SMPL-X repository~\cite{SMPL-X:2019}. 
% Image rasterization is done with CUDA kernels from the official 3D-GS~\cite{kerbl3Dgaussians} implementation. 
We optimize the total training objectives using Adam optimizer with hyperparameters $\beta_1=0.9$ and $\beta_2=0.99$. More implementation details are given in the supplementary materials.
%The training stage takes about 5 hours for 500K iterations on a single Telsla V100 GPU. In the inference stage, our approach can drive and render the animated neural human avatar in real-time (20 FPS). 

\subsection{Dataset and evaluation metrics}

We validate our method on 3 datasets: THuman4~\cite{SLRF} is captured by 24 calibrated RGB cameras at 30 fps with an image resolution of 1330$\times$1150. 3 different subjects are covered, each sequence ranging from 2500 to 5000 frames. DNA-Rendering~\cite{cheng2023dna} uses 60 cameras to capture a wide range of human motions and clothing. ZJU-Mocap~\cite{peng2021neural} is obtained using 23 hardware-synchronized cameras, with 1024$\times$1024 resolution. Similar to existing work, we utilize PSNR, SSIM, and LPIPS to evaluate both novel view and novel pose synthesis. Following PoseVocab~\cite{li2023posevocab}, we also include FID metric~\cite{heusel2017gans} on the THuman4 dataset to measure the realism between rendered and ground truth data.
 
\subsection{Comparison with state-of-the-art}
\label{sec:comparison}

\paragraph{Baselines}

We compare our approach with state-of-the-art methods: 1) PoseVocab~\cite{li2023posevocab} uses SDF-based volume rendering and joint-structured embeddings, 2) SLRF~\cite{SLRF} defines hundreds of local NeRFs defined on SMPL surface, 3)~TAVA~\cite{li2022tava} is a template-free NeRF approach with forward skinning weights field, 4) Ani-NeRF~\cite{peng2021animatable}, uses a canonical NeRF with backward skinning, 5) ARAH~\cite{ARAH:ECCV:2022} defines a canonical SDF and finds backward correspondences with joint root-finding and 6) DVA~\cite{remelli2022drivable} uses forward deformation of articulated volumetric primitives.

%We reproduce novel pose synthesis multi-view experiments from previous work. It includes benchmarking our algorithm on 2 datasets and comparison with a wide range of the most recent prior work on the topic.

\subsubsection{Evaluation on THuman4 dataset}

\begin{figure*}[t]
   \centering
   \includegraphics[width=\linewidth]{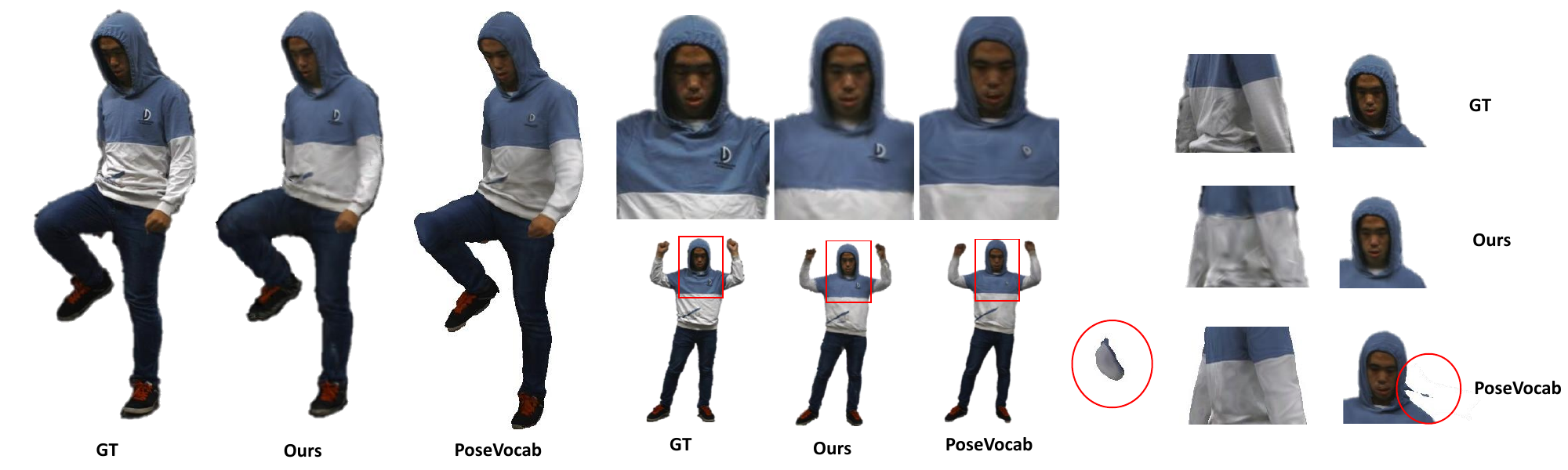}
   \caption{\textbf{Qualitative comparison between PoseVocab and HuGS on Thuman4 dataset.} On the left, our avatar shows better fidelity w.r.t. the body pose than PoseVocab that fails to deform the hood and the knee at the correct location. In the middle, HuGS presents a more detailed logo and more accurate head pose. On the right, PoseVocab presents artefacts due to failures of the inverse skinning process.}
   \label{fig:qualitative}
\end{figure*}

For this dataset, we carefully replicate experiments proposed by PoseVocab~\cite{li2023posevocab} on the ``subject00'' sequence. One of the 24 cameras is held out for the evaluation of novel view synthesis. For novel pose synthesis, the method is trained with the first 2000 frames and evaluated on the rest 500 frames. Quantitative results are given in Tab.~\ref{tab:thuman_quantitative}, where the score of other methods is reported from PoseVocab~\cite{li2023posevocab} and the efficiency metrics have been collected on each respective paper, except for TAVA that does not report rendering time and indicates training time as a limitation. Our method obtains better PSNR, SSIM, and FID scores than all the competitors. In terms of time efficiency, the training time of our method is on par with or better with the baselines, while being the only method to present a real-time rendering in the inference stage. This comparison shows that our method can achieve state-of-the-art performance on both novel view synthesis and novel pose animation while maintaining real-time rendering speed. We show a qualitative comparison against the best competitor PoseVocab in Figure~\ref{fig:qualitative}. HuGS and PoseVocab exhibit different strengths and weaknesses. Our method shows more fidelity to the groundtruth image, for example the logo on the hoody or the head pose which is perfectly aligned. On the other hand, PoseVocab exhibits a smoother surface thanks to SDF formulation. Another drawback of PoseVocab are artefacts that appears outside of the body topology, due to failure in the inverse skinning process. Because we use forward deformation, such artefacts do not happen with our method.

\subsubsection{Evaluation on DNA-Rendering dataset}

\begin{figure*}[t]
   \centering
   \includegraphics[width=\linewidth]{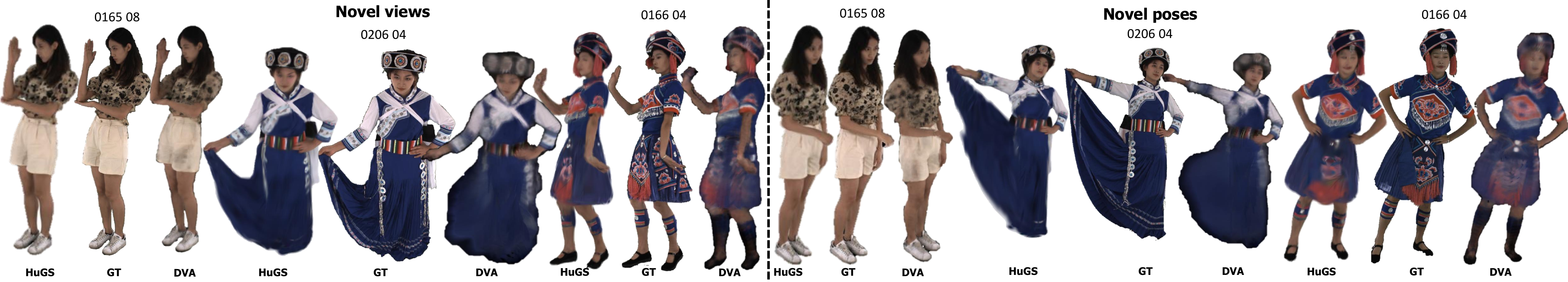}
   \caption{\textbf{Comparison with DVA on the DNA-Rendering dataset.} Despite fast non-rigid motion of complex textured garments, our method preserves more details than DVA and is able to fit unusual topology with loose clothing.}
   \label{fig:dna_qualitative}
\end{figure*}

We further evaluate our method on DNA-Rendering~\cite{cheng2023dna} that proposes more challenging subjects with loose clothing and complex textures. We use 48 cameras and 80 \% of the video to train on 3 sequences. Novel view synthesis is evaluated on 12 held out cameras and novel poses on the held out last frames with all cameras. We compare HuGS to DVA~\cite{remelli2022drivable}, which also performs forward deformation of 3D primitives. Quantitative comparison is provided in Tab.~\ref{tab:dna} and renderings are shown in Fig.~\ref{fig:dna_qualitative}.  The combination of complex textures, fast non-rigid motion and loose clothing make it very difficult for both methods to render details with high fidelity. Nonetheless, our method exhibits better results than DVA and can fit unusual topology and preserve it under novel poses. This is because we rely on the template only at initialization and are able to fit the canonical representation to an arbitrary shape, unlike other methods such as DVA that define primitives close to the template surface.

\input{tables/dna2}

\subsubsection{Evaluation on ZJU-MoCap dataset}

\begin{figure}[b]
   \centering
   \includegraphics[width=\linewidth]{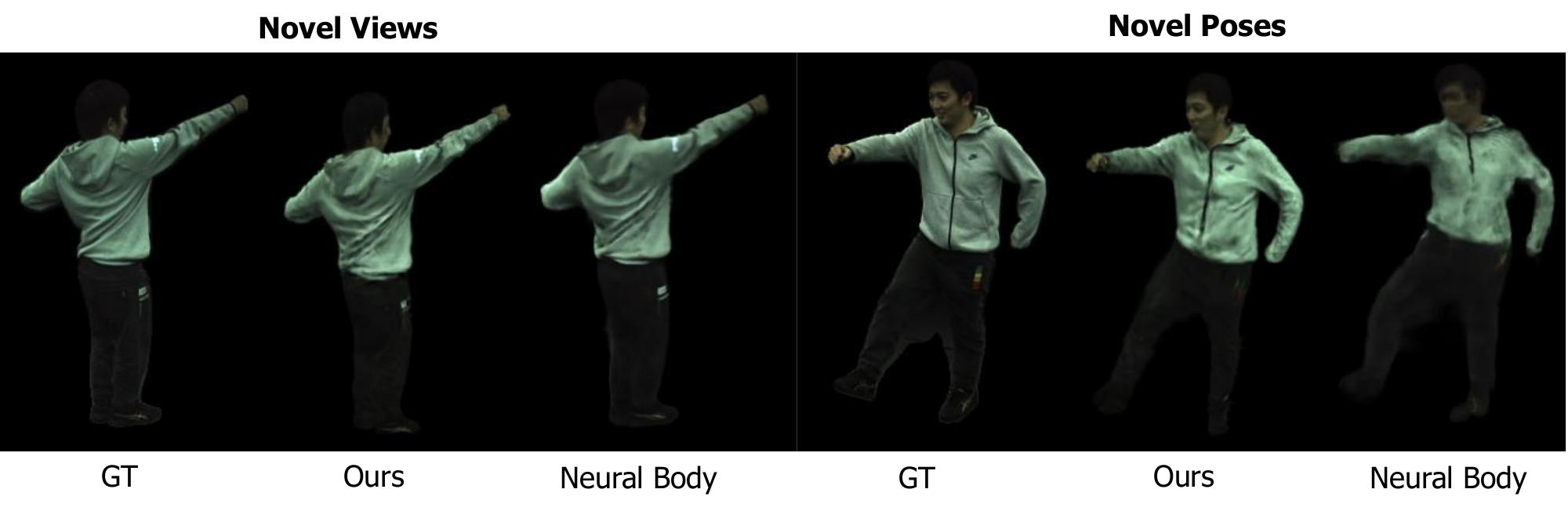}
   \caption{\textbf{Comparison with Neural Body on ZJU-MoCap dataset.} While results in novel pose synthesis are comparable for both methods, HuGS generalizes way better to novel poses thanks to its forward deformation formulation.}
   \label{fig:zju_figure}
\end{figure}

To benchmark HuGS on ZJU-MoCap~\cite{peng2021neural}, we follow the setting of NeuralBody~\cite{peng2021neural}, i.e. use only 4 out of 21 cameras to train, and report results given by SLRF~\cite{SLRF}. On this dataset, we train only for 50k iterations because we observe that textures degrade on novel poses with further training, due to the sparse camera setup. Table~\ref{tab:zju} presents the quantitative results. Our method achieves the second-best performance on novel views and the best performance on novel pose synthesis, where we obtain the highest PSNR and competitive SSIM. Moreover, our approach is notably more efficient for training and rendering compared to the baselines. These results shows that HuGS can generalize well in novel pose animation tasks, as shown in Fig.~\ref{fig:zju_figure}.

\input{tables/zju_387}

\subsection{Ablation studies}
\label{sec:ablations}

\begin{figure}[b]
   \centering
   \includegraphics[width=\linewidth]{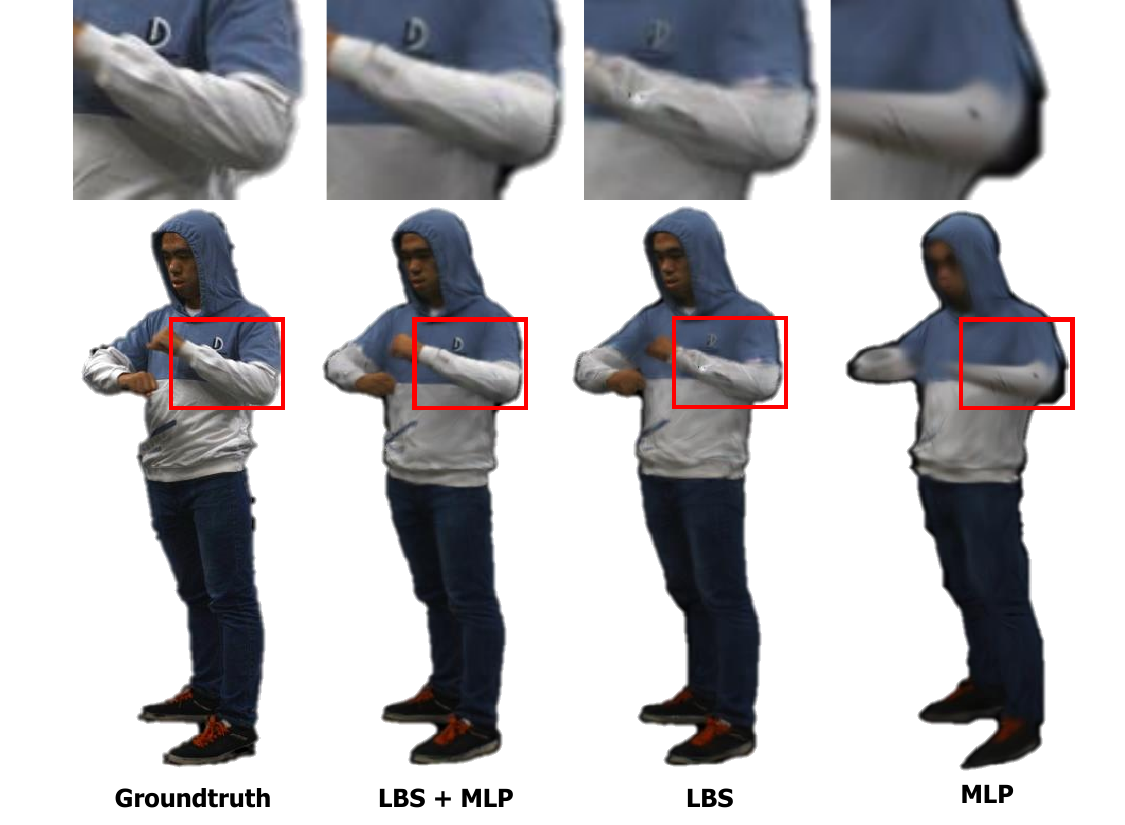}
   \caption{\textbf{Ablation study on the importance of combining LBS and MLP.} We show the images in the order of ground truth, qualitative results of our full model with both LBS and posed-based MLP, and the ablated model with LBS only and with MLP only. }
   \label{fig:ablation_lbs_mlp}
\end{figure}

\paragraph{Animate with a single transformation} The core of our method consists in combining two transformations: LBS and a pose-dependent MLP. We study the performance of our method by using either LBS or MLP as the only deformation. Qualitative are shown in Fig.~\ref{fig:ablation_lbs_mlp}. Using only LBS is a strong baseline quantitatively competitive with the state-of-the-art. However, because it only encodes the rigid deformations of body joints, it is not able to model wrinkles on garments and lacks details compared to the proposed method. The MLP only version fails to model large deformations such that arms are not represented and thus is not a suitable algorithm for animatable avatars.

\paragraph{Learning skinning weights} One contribution of our work is to learn per-gaussian skinning weights. We compare this design choice to a simpler baseline: a model where weights are not learned but imported from the closest SMPL vertex. Because the MLP can potentially correct errors from previous steps, we deactivate it for this experiment and deform gaussians only with LBS. We present the comparison in Tab.~\ref{tab:ablation_skinning}, where we observe that the model with learned skinning weights can bring more than 1dB improvement on PSNR over the SMPL skinning weights. The main reason is that SMPL weights are defined on the naked body while ours can adapt freely to any body shape. Template weights could also be inaccurate due to the pose estimation error in the training data, causing misalignement between the canonical gaussians and the template mesh.

\input{tables/skinning_weights_ablation}

\paragraph{Latent code or position} We use a per-gaussian latent code as input of our MLP. It is in contrast with previous work~\cite{yang2023deformable3dgs} that used the position to identify the primitive. We train a model where we replace the latent code (16 floats) by canonical position augmented with fourier features~\cite{tancik2020fourier} (63 floats). As shown in Tab.~\ref{tab:ablation_position}, latent codes perform slightly better than position while being more compact. With more MLP layers, we expect positional encodings to work similarly, but the learned features help to decode the information in small MLPs, similar to explicit features grids that accelerate NeRFs~\cite{mueller2022instant}.

\input{tables/position_ablation}

\paragraph{Shading} Finally, we verify the benefit of the shading component of our pipeline. We train a model where the MLP only outputs translation and rotation for each gaussian on the sequence00 of THuman4 dataset. This model obtains the following metrics on novel pose synthesis: 29.72dB PSNR, 0.978 SSIM, 0.027 LPIPS. These scores are directly comparable with those displayed in Tab~\ref{tab:thuman_quantitative}. Removing this component forces the model to learn duplicate gaussians with different colors for shaded areas, leading to overfitting.

\subsection{Efficiency}

One of the main advantage of our method against previous work is its rendering time.
We render an image of size 512x512 in $\approx 12$ms or $80$fps on a single Tesla V100 GPU. This runtime is two orders of magnitude faster than the compared SoTA, as shown in Table~\ref{tab:thuman_quantitative}. Computing transformations from canonical to observation space takes $10$ms and gaussians rasterization $2$ms.
%Moreover, for many applications of animatable avatars, this time can be greatly reduced by pre-computing transformations offline before rendering. This is the case when the avatar only needs to be rendered with a pre-defined set of gestures. In this scenario, each body pose can be stored as a static 3D-GS model and novel views can be rendered in potentially 2ms. Our avatar models contains usually around 40k gaussians, that corresponds to a memory footprint of $2.2$MB per frame that need to be cached in memory.
Training the model for our experiments takes from 5 to 20 hours on a Tesla V100, depending on the dataset size. It compares equally or favorably to neural rendering competitors.

%% file: tables/thuman_full.tex
\begin{table*}[t]
    \small
    \centering
    \caption{\textbf{Quantitative comparison on Thuman4 dataset.} We evaluate the performance on both novel view and novel pose synthesis, and time efficiency. Our method achieves the best performance on all the metrics and supports real-time rendering in the inference stage.}
    \label{tab:thuman_quantitative}
    \begin{tabular}{lcccc|cccc|cc}
    \hline
    \multirow{2}{*}{\textbf{Method}} & \multicolumn{4}{c}{Training poses} & \multicolumn{4}{c}{Novel poses} & \multicolumn{2}{c}{Efficiency} \\
    \cline{2-11}
     & \textbf{PSNR} & \textbf{SSIM} & \textbf{LPIPS} & \textbf{FID} & \textbf{PSNR} & \textbf{SSIM} & \textbf{LPIPS} & \textbf{FID} & \textbf{Render (s)} & \textbf{Training (h)} \\
    \hline
    \textbf{\OURS (Ours)} & \cellcolor{orange}35.05 & \cellcolor{orange}0.99 & 0.020 & \cellcolor{orange}9.48 & \cellcolor{orange}32.49 & \cellcolor{orange}0.984 & 0.019 & \cellcolor{orange}11.76 & \cellcolor{orange}0.015 & \cellcolor{orange}10 \\
    PoseVocab~\cite{li2023posevocab} & 34.23 & 0.99 & \cellcolor{orange}0.014 & 23.957 & 30.97 & 0.977 & \cellcolor{orange}0.017 & 37.239 & 3 & 48+ \\
    SLRF~\cite{SLRF} & 25.27 & 0.97 & 0.024 & 44.49 & 26.15 & 0.969 & 0.024 & 110.651 & 5 & 25 \\
    TAVA~\cite{li2022tava}  & 23.93 & 0.97 & 0.029 & 75.46 & 26.61 & 0.968 & 0.032 & 99.947 & - & - \\
    Ani-NeRF~\cite{peng2021animatable} & 23.19 & 0.97 & 0.033 & 85.45 & 22.53 & 0.964 & 0.034 & 102.233 & 1.09 & 12 \\
    ARAH~\cite{ARAH:ECCV:2022} & 22.02 & 0.96 & 0.033 & 74.30 & 21.77 & 0.958 & 0.037 & 77.840 & 10 & 36 \\
    \hline
    \end{tabular}
    
\end{table*}

%% file: tables/dna2.tex
\begin{table}[h]
    \footnotesize
    \centering
    \caption{\textbf{Comparison with DVA on DNA-Rendering.}}
    \label{tab:dna}
    \begin{tabular}{llccc|ccc}
    \hline
    \multirow{2}{*}{\textbf{Seq}} & \multirow{2}{*}{\textbf{Method}} & \multicolumn{3}{c}{Novel views} & \multicolumn{3}{c}{Novel poses}\\
    %\cline{2-11}
     & & \textbf{\tiny{PSNR}} & \textbf{\tiny{SSIM}} & \textbf{\tiny{LPIPS}} & \textbf{\tiny{PSNR}} & \textbf{\tiny{SSIM}} & \textbf{\tiny{LPIPS}} \\
    \hline
    \multirow{2}{*}{0165} & HuGS & \cellcolor{orange}31.5 & \cellcolor{orange}0.98 & \cellcolor{orange}0.022 & \cellcolor{orange}30.0 & \cellcolor{orange}0.97 & \cellcolor{orange}0.025 \\
     & DVA~\cite{remelli2022drivable} & 29.8 & 0.97 & 0.025 & 28.8 & 0.97 & 0.036 \\
     \hline
     \multirow{2}{*}{0166} & HuGS & \cellcolor{orange}27.0 & \cellcolor{orange}0.97 & \cellcolor{orange}0.050 & \cellcolor{orange}25.7 & \cellcolor{orange}0.96 & \cellcolor{orange}0.056 \\
     & DVA~\cite{remelli2022drivable} & 26.1 & 0.96 & 0.059 & 25.4 & 0.95 & 0.063 \\
     \hline
     \multirow{2}{*}{0206} & HuGS & \cellcolor{orange}25.7 & \cellcolor{orange}0.96 & \cellcolor{orange}0.061 & \cellcolor{orange}23.2 & \cellcolor{orange}0.94 & \cellcolor{orange}0.073 \\
     & DVA~\cite{remelli2022drivable} & 22.8 & 0.94 & 0.076 & 23.1 & 0.93 & 0.079 \\
    \hline
    \end{tabular}
    
\end{table}

%% file: tables/zju_387.tex
\begin{table}[h]
    \centering
    \caption{\textbf{Results on ZJU-MoCap dataset.} We present the second-best performance on novel view synthesis while outperforming all the baselines on PSNR metric for novel poses.}
    \label{tab:zju}
    \begin{tabular}{lcccc}
    \hline
    & \multicolumn{2}{c}{Novel Views} & \multicolumn{2}{c}{Novel Poses} \\
    \hline
    \textbf{Method} & \textbf{PSNR} & \textbf{SSIM} & \textbf{PSNR} & \textbf{SSIM}\\
    \hline
    \textbf{\OURS (Ours)} & 26.58 & 0.934 & \cellcolor{orange}23.69 & 0.896 \\
    SLRF~\cite{SLRF} & \cellcolor{orange}28.32 & \cellcolor{orange}0.953 & 23.61 & \cellcolor{orange}0.905 \\
    Neural Body~\cite{peng2021neural} & 25.79 & 0.928 & 21.60 & 0.870 \\
    Ani-NeRF~\cite{peng2021animatable} & 24.38 & 0.903 & 21.29 & 0.860 \\
    \hline
    \end{tabular}
    
\end{table}

%% file: tables/skinning_weights_ablation.tex
\begin{table}[h!]
    \centering
    \caption{\textbf{Ablation study on skinning weights.} We evaluated the ablated model with learned skinning weights and weights from the SMPL template on novel view synthesis. }
    \label{tab:ablation_skinning}
    \begin{tabular}{lcccc}
    \hline
    \textbf{Method} & \textbf{PSNR} & \textbf{SSIM} & \textbf{LPIPS} & \textbf{FID} \\
    \hline
    Learned & 31.99 & 0.984 & 0.020 & 27.15 \\
    Template & 30.97 & 0.981 & 0.022 & 28.50 \\
    
    \hline
    \end{tabular}
    
\end{table}

%% file: tables/position_ablation.tex
\begin{table}[h]
    \centering
    \caption{\textbf{Ablation study on MLP input}. We compare learnable latent codes with canonical position encoding as MLP input, for novel pose synthesis on THuman4 Dataset.}
    \label{tab:ablation_position}
    \begin{tabular}{lcccc}
    \hline
    \textbf{Method} & \textbf{PSNR} & \textbf{SSIM} & \textbf{LPIPS} & \textbf{FID} \\
    \hline
    Latent code & 32.49 & 0.984 & 0.019 & 11.76 \\
    Position & 32.20 & 0.985 & 0.019 & 17.78 \\
    
    \hline
    \end{tabular}
    
\end{table}

%% file: sec/5_discussion.tex
\section{Discussion}
\label{sec:discussion}

\paragraph{Limitations and future work} 
Our method is limited in several aspects: 1) the 3D reconstruction quality degrades with sparse camera setups because of overfitting on the training observations, resulting in poor novel pose synthesis. 2) The proposed MLP is able to fit observed garment deformations and replicate them for novel poses, but does not extrapolate novel deformations. 3) Each gaussian in our model is optimized and deformed independently, ignoring the relation between gaussians in local neighbourhoods. We think that defining structure and connectivity between primitives would help and leave it as future work. 

\paragraph{Potential societal impacts} The proposed algorithm could be used in Deep Fakes pipelines to synthesize fake videos of people with reenactment for malicious purpose. This aspect needs to be addressed and mitigated carefully.

\paragraph{Concurrent works} There is a wide-spread interest in human modelling with gaussian splatting, such that many preprints with related approaches have been released during the review process of this paper~\citep{zielonka2023drivable,jena2023splatarmor,li2023animatable,lei2023gart, dhamo2023headgas,liu2023animatable,kocabas2023hugs, pang2023ash, qian20233dgs, li2023human101, jung2023deformable, li2024gaussianbody, hu2023gaussianavatar, jiang2023hifi4g, qian2023gaussianavatars}. We discuss key similarities and differences in the supplementary materials.

%% file: sec/6_conclusion.tex
\section{Conclusion}
\label{sec:conclusion}

We have proposed \textbf{\OURS}, a first approach for creating and animating virtual human avatars based on Gaussian Splatting, by defining a coarse-to-fine deformation algorithm that combines forward skinning with local learning-based refinement.  
Using Gaussian Splatting for this problem not only helps to accelerate rendering, but also enables to bypass difficult inverse skinning approaches required in NeRF-based formulations. In contrast with other forward approaches, we are able to fit body shapes with loose clothing. Experimental results demonstrate that our approach can achieve state-of-the-art human neural rendering performance with good generalization. The fast rendering of this approach should facilitate its deployment and we hope that HuGS will also serve as an intuitive baseline for follow-up research on gaussian-based avatars.

%% file: sec/X_suppl.tex
\clearpage
\setcounter{page}{1}
\maketitlesupplementary

We present further analysis of our method. We invite readers to watch the video that summarizes our contributions and demonstrates real-time rendering, whose details are given in section~\ref{sec:video}.
We further present extensive implementation details and hyperparameters setting in section~\ref{sec:reproducibility}, in order to facilitate the reproducibility of our experiments. Finally, we provide more qualitative results of our method on the THuman4 dataset in section~\ref{sec:quali}. 

 \begin{figure*}[h!]
    \centering
    \includegraphics[width=\linewidth]{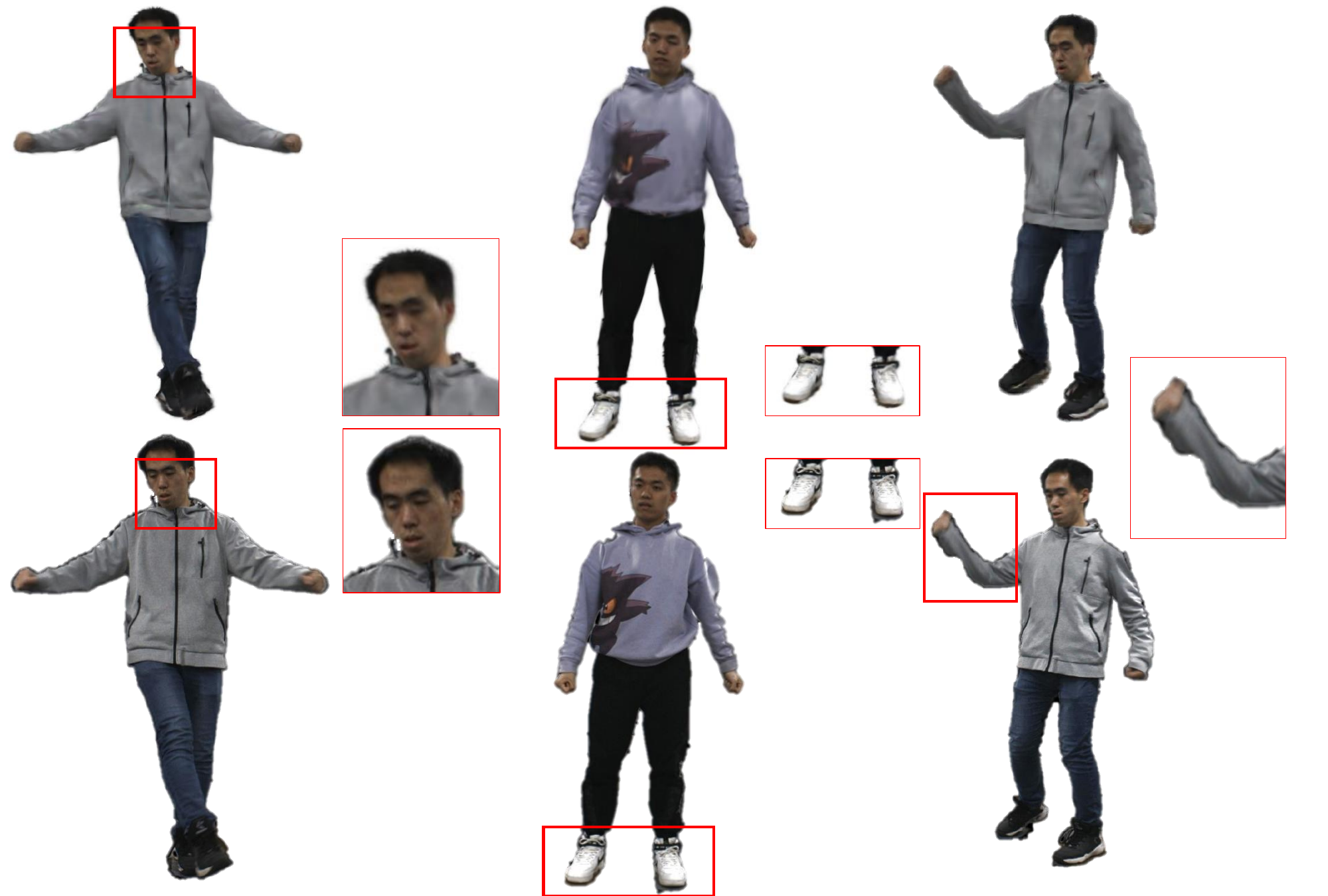}
    \caption{\textbf{Qualitative visualization of HuGS novel pose synthesis on THuman4 dataset.}}
    \label{fig:quali}
 \end{figure*}

\section{Real-time video rendering}
\label{sec:video}
\paragraph{Real-time rendering in the supplementary video} The attached video showcases real-time novel pose synthesis on the THuman4 dataset~\cite{SLRF} at 60fps. This is done by extending the viewer from 3D-GS to dynamic scenarios in order to render videos.

\section{Reproducibility details}
\label{sec:reproducibility}

We have described our main implementation details in the main manuscript. In this section, we further report the hyperparameter values of our pipeline in Table~\ref{tab:hyperparameters}. After that, we provide the additional implementation details of our approach, as described as below. 

\paragraph{Linear blend skinning} Our implementation for linear blend skinning (LBS) follows SMPL-X~\cite{SMPL-X:2019}. Notably, we do not use pose blend shapes before applying deformations on human joints, because the shape estimation and blend shapes obtained upon that may be inaccurate, thus we design MLP to handle pose-dependent deformations. We also remind that in our case, LBS is applied on canonical gaussians only and thus deforming template vertices is not necessary. The learnable per-gaussian skinning weights vector $\mathbf{w}$ is a parameter optimized through gradient descent which can leads to negative values. We apply a ReLU activation on each $\mathbf{w}_{j}$ and then normalize the vector such that its components sum to 1 to obtain a well defined skinning weights vector. Finally, the transformations matrices $\mathbf{M}_{j,t}$ that encode the rigid deformation of each body joint $j$ for each training timestep $t$ are precomputed before training to improve efficiency.

\paragraph{Learning rates} Similar to 3D-GS~\cite{kerbl3Dgaussians}, we use different learning rates for each set of learnable parameters. We leave the learning rates of the original parameters (position, orientation, scaling, colors and opacity) unchanged. Our MLP and the skinning weights vectors $\mathbf{w}$ are optimized with a constant learning rate that is set to $1e^{-4}$. For latent codes $\mathbf{l}$, we use a learning rate of $2.5e^{-3}$.

\begin{table}[h]
    \centering
    \caption{\textbf{Hyperparameters values.}}
    \label{tab:hyperparameters}
    \begin{tabular}{lc}
    \hline
    \textbf{Parameter name} & \textbf{Value} \\
    \hline
    $\lambda_{L_1}$ (in Eqn. 8) & 0.8 \\
    $\lambda_{\text{ssim}}$ (in Eqn. 8) & 0.2 \\
    $\lambda_{\text{lpips}}$ (in Eqn. 8) & 0.05 \\
    $\lambda_{\text{trans}}$ (in Eqn. 8) & 0.01 \\
    $\lambda_{\text{rot}}$ (in Eqn. 8) & 0.001 \\
    $\lambda_{\text{s}}$ (in Eqn. 8) & 0.001 \\
    $\lambda_{\text{mesh}}$ (in Eqn. 8) & 0.1 \\
    $\lambda_{\text{skn}}$ (in Eqn. 8) & 0.001 \\
    Dimension of latent code $\mathbf{l}$ & 16 \\
    
    \hline
\end{tabular}
    
\end{table}

\section{Concurrent works}
\label{sec:concurrent}

Gaussian splatting is currently a very active research topic and while this work was under review, many related drivable avatars based on gaussian splatting have been released. We propose a small discussion and refer to \href{https://github.com/MrNeRF/awesome-3D-gaussian-splatting}{Awesome 3D Gaussian splatting} for a complete list of related papers.

Similar to HuGS, most gaussian avatars exploit forward deformation of gaussians with LBS to drive the avatar. Local refinement with a neural network is also a popular choice~\citep{jena2023splatarmor, li2023animatable, li2023human101, jung2023deformable} but different designs have been developed. Notably, SplatArmor~\citep{jena2023splatarmor} uses canonical gaussian parameters as input, Animatable Gaussians~\citep{li2023animatable} and ASH~\citep{pang2023ash} use a 2D CNN. The per-gaussian latent code and the shading components proposed by our method are the main advantages of HuGS compared to these approaches. Regarding skinning weights, most methods rely on the template, with the exception of GART~\citep{lei2023gart} that also optimizes these parameters. In contrast with ours, these learnable skinning weights are not defined per-gaussian but in a voxel grid. Finally, similar to ours, most methods optimize the canonical gaussians jointly with the rest of the pipeline. In contrast, Animatable Gaussians~\citep{li2023animatable} and ASH~\citep{pang2023ash} import a template (such as a SDF) where the body shape as already been fitted on the subject. This design choice adds an expensive pre-processing step but also seems to exhibit very good results. We expect follow-up research to build on this large amount of proposals to push gaussian-based animatable avatars forward.

\section{Qualitative analysis}
\label{sec:quali}

We display in Figure~\ref{fig:quali} qualitative results of the HuGS method for subject01 and subject02 sequences from the THuman4 dataset~\cite{SLRF} for novel pose synthesis. Note that no quantitative comparison is done on these subjects because the evaluation setup has not been released by the dataset authors. We observe that our method is able to fit the subjects with precise details, such as the black hood button (left picture) or the shoes (middle), and render the target body pose with high fidelity. However, we also showcase inaccuracies in the dataset caused by segmentation masks and motion blur that are observed regularly on training images and thus create artifacts in the learned model and degrade the overall rendering quality on these subjects.